\documentclass[publish,JACIII,paper]{jaciiiarticle}
\usepackage[dvips]{graphicx} 

\usepackage{amssymb}

\usepackage{amsmath}
\allowdisplaybreaks[0]
\usepackage{graphicx}
\usepackage{epsfig}
\usepackage{mathrsfs}
\usepackage{subfigure}
\usepackage{booktabs}
\usepackage{cite}

\usepackage[english]{babel}
\usepackage{algorithm}
\usepackage{algorithmic}

\SetVolumeNumber{Vol.0 No.0, 200x} 
\SetArticleID{jc*-**-**-****:}

\begin{document}

\pagestyle{jaciiistyle}

\title{Generative Adversarial Network based on Resnet for Conditional Image Restoration}
\author{Meng Wang, Huafeng Li, Fang Li}
\address{Kunming University of Institute Technology, Information and Automatic College\\
         E-mail: vicong68@qq.com}
\markboth{Wang Meng}{Generative adversarial network based on resnet for conditional image restoration}
\dates{00/00/00}{00/00/00}
\maketitle

\begin{abstract}
\noindent The GANs promote an adversarive game to approximate complex and jointed example probability.
The networks driven by noise generate fake examples to approximate realistic data distributions.
Later the conditional GAN merges prior-conditions as input in order to transfer attribute vectors to the corresponding data. 
However, the CGAN is not designed to deal with the high dimension conditions since indirect guide of the learning is inefficiency.
In this paper, we proposed a network ResGAN to generate fine images in terms of extremely degenerated images.
The coarse images aligned to attributes are embedded as the generator inputs and classifier labels. 
In generative network, a straight path similar to the Resnet is cohered to directly transfer the coarse images to the higher layers.
And adversarial training is circularly implemented to prevent degeneration of the generated images. 
Experimental results of applying the ResGAN to datasets MNIST, CIFAR10/100 and CELEBA show its higher accuracy to the state-of-art GANs.
\end{abstract}


\section{Introductions}
To generate realistic images, networks with deep layers proposed recently have been proved beyond the other shallow approaches \cite{Vincent2008,Radford2015}.   
In the generative networks, 
the generative adversarial networks (GANs) proposed in \cite{Goodfellow2014} implement an alternated training scheme motivated by adversarial game,
i.e. the generator $G$ and discriminator $D$ attack each other to benefit itself.
If the game reaches a balance condition, 
the generator has learned to generate examples that approximate realistic data distributions.
Instead of directly estimating probabilities, 
GANs promote an adversarial training scheme to solve the generation for high dimension and complex distributions,
and therefore are successfully implemented to generate images and videos freely or conditionally \cite{Mirza2014,yu2017seqgan}.   

The original GAN motivated by the success of neural networks implementations such as CNN \cite{Simonyan2014} and Auto-encoder \cite{Vincent2008}. 
In \cite{Radford2015} the authors proposed deep convolution GAN (DCGAN) that aligns 2 deep networks to generate more quality adversarial examples 
such as images \cite{Zhu}, text \cite{Reed2016} or video sequences \cite{yu2017seqgan}.
The deep networks have abundant filter banks to promote more detailed analysis and synthesis of large examples.
These convolution layers are adjusted by the propagated gradients of cost functions.
The cost functions drawn to the GANs and DCGANs are logistic type, such as sigmoid and softmax functions. 
Recently, in \cite{Gan2017} the authors implements Wasserstein distance as an efficient measurement between the fake and the real images, 
and the scheme has been verified more robustness and efficient.

According to the GANs or DCGANs, the generative network driven by noise of uniform distribution has no conditional representation to assign.
To generate conditionally, 
the CGAN merges determined conditional variables as inputs to transfer all the classes and attributes to the corresponding examples \cite{Mirza2014}. 
The modalities of CGAN are more easy to learn as the conditional distributions $P(X|Y)$ are simpler than the rough margin distributions $P(X)$.
However, the CGAN is not a discriminated method since the conditions as input variables cannot directly guide the weights learning.
Therefore conditional semantics are hard to be revealed with manually selecting the random inputs vectors.
Moreover, the CGANs formulate the outputs as single binary labels with bit probabilities (Bernoullis), 
and which implies that these networks cannot transfer the original distribution to abundant semantic distributions.

On the other hand, the deep networks have lots of paths to learn the detailed features.
But the original feature representations are easy to diffuse during pass through layers. 
And the generated results thus go into under-fitting or local maximization.
To overcome it, the authors in \cite{Windows2014} claimed that the batch normalization enhances the robustness of representation learning,
and is then implemented in DCGAN with deeper networks \cite{Radford2015}.
We notice that the Resnet structure proposed in \cite{He2015} also make the features go deeper in network,
and to our knowledge the Resnet have not been found to implemented in generative networks, i.e. GANs. 

To restore images, the coarse images can be adopted as the conditional inputs in the generations.
The proposed network is similar to CGAN but a classifier is embedded into the discriminator,
so that the classified labels can be no longer as the input for both modules $G$ and $D$.
Also the pass-through channel in Resnet \cite{He2015} is not only useful to classifications, 
but we can push the coarse images to the pass-through channel to maintain the original spatial features.
Based on them we finally formulate a uniform and efficient image generation network.

In this paper, a ResGAN model is proposed for the task of image restoration.
We extend the traditional GAN schemes in terms of directly using coarse images features and the attributes labels. 
Therefore, the input images are efficiently restored in terms of discriminative features learned by the iterations of adversarial training. 
Also a classifier is embedded to the discriminator thus the performance of generations is enhanced and the adversarial training is more robustness.
The experimental results evaluated by loss and accuracy verified the proposed model is superior to the state-of-art GANs.

The structure of the following sections is illuminated.
In Section 2, the state-of-art GANs are summarized,
also the proposed ResGAN model are revealed.
The experimental results and discussions are given in Section 3.
The last section summarizes this paper.

\section{Methods}
\subsection{GANs review}
According to the original GANs, the training process is considered as a game between the discriminator $D$ and the generator $G$,
and the objective of the adversary between the 2 networks can be integrated into a uniform equation
\begin{equation}
 \begin{aligned}
   \min\limits_{G}\max\limits_{D} V(D,G)=&\mathbb E_{x\sim p_{\rm data}(x)}[\log D(x)]+ \\
   &\mathbb E_{z\sim p_z(z)}[\log (1 - D(G(z)))].
\end{aligned}
\end{equation}
which is a max-minimizing process with adversarial iterations to obtain parameters of both the modules.  
The adversarial iterations will continue to approximate the theoretic balance solution,   
and finally learn the modular parameters that can accurately generate or discriminate the distribution of the data $x$. 
Another version of GAN, i.e. the WGAN, employs the adversarial learning on different loss measurement 
\begin{equation}
 \begin{aligned}
   \min\limits_{G}\max\limits_{D} V(D,G)=&\mathbb E_{x\sim p_{\rm data}(x)}D(x)- \\
   &\mathbb E_{z\sim p_z(z)}D(G(z)).
\end{aligned}
\end{equation}
and the final output is then obtained based on minimizing the simplified Wasserstein distance of the fake and the real examples. 
The WGAN have enhanced generative performances without the cost of the additional calculations.

By the assumption that the distribution of data $x$ is the margin of the conditional distribution $P(y)$, 
the representation of condition $y$ is then implicitly embedded to the conditional probability $p(x|y)$.
According to this, the object equation of the CGAN is written as 
\begin{equation}
 \begin{aligned}
   \min\limits_{G}\max\limits_{D} V(D,G)=&\mathbb E_{x\sim p_{\rm data}(x)}[\log D(x|y)]+ \\
   &\mathbb E_{z\sim p_z(z)}[\log (1 - D(G(z|y)))].
\end{aligned}
\end{equation}
where the condition $y$ is the control input to both the modules $D$ and $G$. 

\subsection{ResGAN model}
\subsubsection{Discriminator by embedding classifier}
The image restoration is a challenge task due to the complex and unknown distributions of real images.
Accroding to the datasets, the algorithms should learn exactly the transfer model between the coarse and fine images.      
In this paper, we introduce a ResGAN scheme based on the original GANs to overcome these hard problems.
The details are focus on how to utilize the example attributes and introduce Resnet to achieve the task.

The proposed ResGAN is formulated by the methodology of maximizing likelihood (ML) as the former GANs.
The difference is that the ResGAN extends the discriminating task from single binary, i.e. true or false, to the multiple attribute representations.   
Therefore, the attribute representation $y$ is utilized as a supervised reference to obtain the maximized negative cross entropy.
Compared to the CGAN, the reference probability $q^0(y)$ is taken apart from the conditional probability $p(.|y)$ to obtain a directly interactive formulation
\begin{equation}
 \begin{aligned}
   \min\limits_{G}\max\limits_{D} V(D,G)=&\mathbb E_{q^0(y)}[\log D(x)]+ \\
   &\mathbb E_{q^0(y)}[\log (1 - D(G(y)))].
\end{aligned}
\end{equation}
The attribute variable $y$ composes the joint distribution $P(Y=(y_1,y_2,...y_d)^{\rm T})$ with independent Bernoulli distribution $B_Y=(B_{y_1},B_{y_2},...B_{y_d})$. 
Base on this, the equation corresponding to multi-attribute $y_k$ is written as
\begin{equation}
 \begin{aligned}
   \min\limits_{G}\max\limits_{D} V(D,G)=&\sum_{k=1}^d\mathbb E_{q^0(y_k)}[\log D_k(x)]+ \\
   &\mathbb E_{q^0(y_k)}[\log (1 - D_k(G_k(y_k)))].
\end{aligned}
\end{equation}
The accumulating to $k$ is included in vector operations and we then remove it for reduction. 
Given the Bernoulli representation $Y$ and the joint function $D(X)$, the object function is then equivalence to
\begin{equation}
 \begin{aligned}
   \min\limits_{G}\max\limits_{D} &V(D,G)=Y\{\log [D(X)(1 - D(G(Y)))]\}
   \\&+(1-Y)\{\log [(1-D(X))D(G(Y))]\}.
\end{aligned}
\end{equation}
Let $J=V(f_\theta(X),g_\phi(Y))$, $f_\theta\triangleq D$ and $g_\phi\triangleq Y$, where $\theta$ and $\phi$ are the parameters to be optimized.
Thus according to the gradient ascent the regulate value of the discriminative parameter $\theta$ is denoted as
\begin{equation}
   \frac{\partial J}{\partial\theta}=(Y-f_\theta(X))X+(1-Y-f_\theta(g_\phi(Y)))Y
\end{equation}
and the generative parameter $\phi$ optimized in terms of the gradient descend as 
\begin{equation}
   \frac{\partial J}{\partial\phi}=(1-Y-f_\theta(g_\phi(Y)))Y
\end{equation}
which has the same formulation only for the second term in the former equation. 
Therefore the adversary learning between these 2 moduals is disequilibrium and tends to obtain dynamic results during the iterative optimization.

\begin{center} {\centering
\vbox{\centerline{\includegraphics[width=10cm,bb=10 13 528 455]
{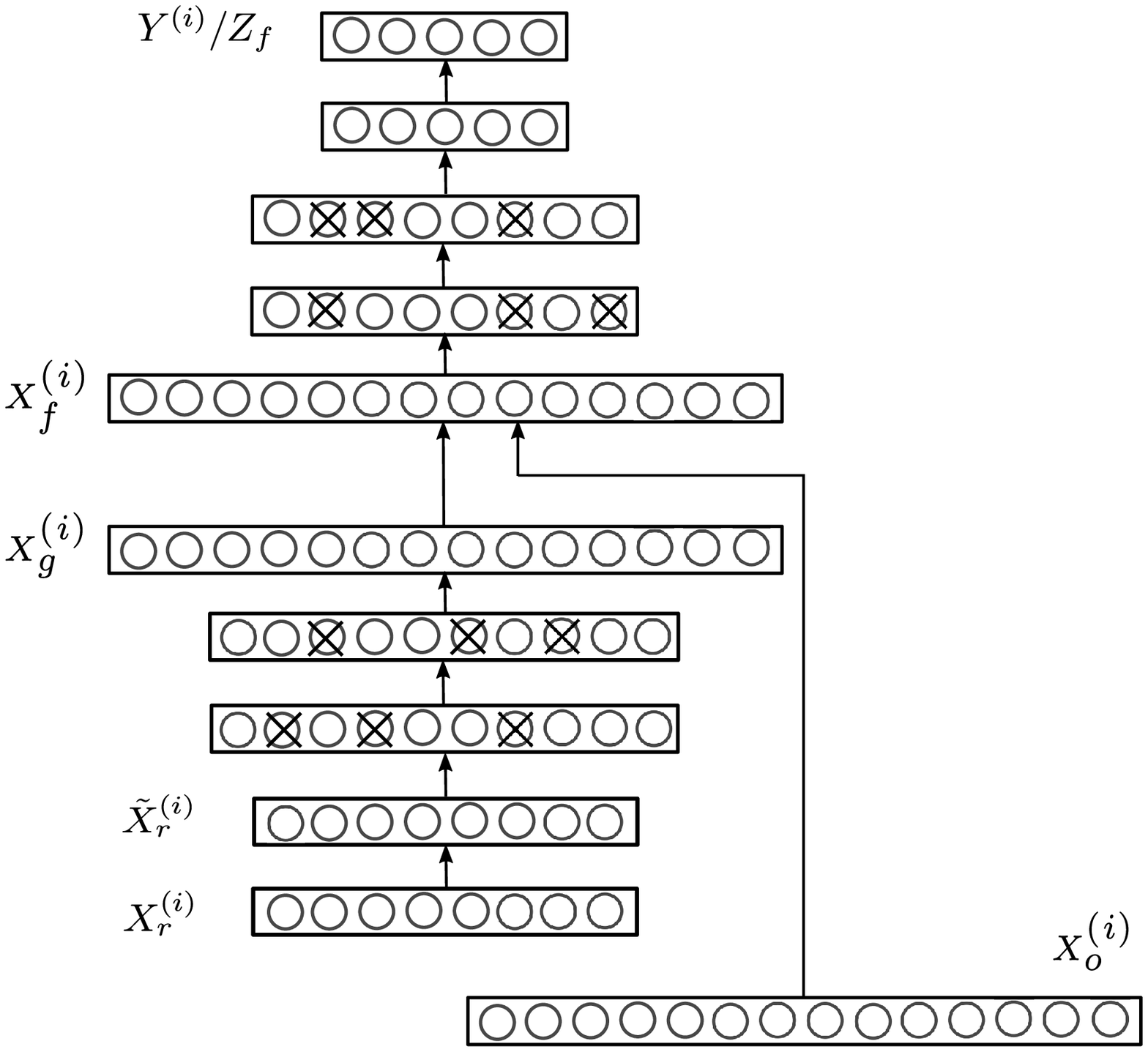}} \vskip1mm {\small
Fig.\,1\quad The architecture of ResGAN with classifier embedding.}}}\end{center}

\subsubsection{Generator with Resnet structure}
The generators of the GANs are formulated as the stacked of deconvolution layers,
and the whole structure as deconvolution networks (DCNNs).
The DCNNs that transfer few features to complex examples are the inverse operations to the CNNs.
And  they have proved to be efficient in realistic image generation.
But in conditional generations, the input conditional variables should be passed across many layers to affect the finally generated outputs,
which is defective especially to the image restoration.

In image restoration, the coarse images contain spatial features that are often aligned to large scale representations.
These spatial alignments are efficient to maintained in image restorations.
We construct a generator with coarse images as input and a Resnet pass-through channel as directly spatial feature transfer. 
In the proposed ResGAN, the coarse images can be directly transferred to the outputs, 
thus the implementations of image repairing are similar to a kind of residual learning.
The repairing process can be denoted as
\begin{equation}
   X_g = f(g(X_r) + X_r)
\end{equation}
where $X_r$ is the coarse image, $X_g$ denotes the fine output 
and $g(X_r)$ denotes the residual image to be learned by the generator.
The generator based on Resnet is shown in Fig.2.
 
\begin{center} {\centering
\vbox{\centerline{\includegraphics[width=8cm,bb=6 11 218 78]
{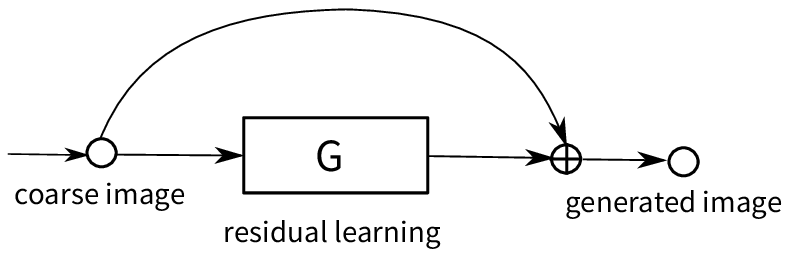}} \vskip1mm {\small
Fig.\,2\quad The basic structure of the generator based on Resnet.}}}\end{center} 

The original Resnet directly adopt the plus operation to the tensors from the pass-through channel and the weighted layers, 
as shown in Eq. 2. Instead of it, we further designed the operation by connecting different tensors passed by the previous network paths into one in a fixed axis,
and then push the connected tensors to the subsequent layers. 
Our version of Resnet can deal with complex tensors across convolution operations to prevent the overwhelm of each other by the adding operation. 

\section{Results}
According to the aboves, the restored images should be similar to the realistic examples, also easily classified by human or machines.
Thus, the written number dataset MNIST is firstly evaluated by both generated image and the classified accuracy.
Fig. 3 shows the generated images with ResGAN generator for the MNIST dataset.
The generated images can be clearly discriminated to different number classes between the rows,
and the handwritten number have various styles that is shown by each in the same rows.
The results denote that the coarse inputs shown as Fig. 5 are transferred to the various distributions of the handwritten styles.
Fig. 4 denotes the background images with the number labels as supervised attributes.
In terms of ResGAN scheme, the classified labels are used to enhance the generative accuracy therefore the fake numbers in Fig. 3 can be discriminated easily.

\begin{center} {\centering
\vbox{\centerline{\includegraphics[width=8cm,bb=0 -1 907 589]
{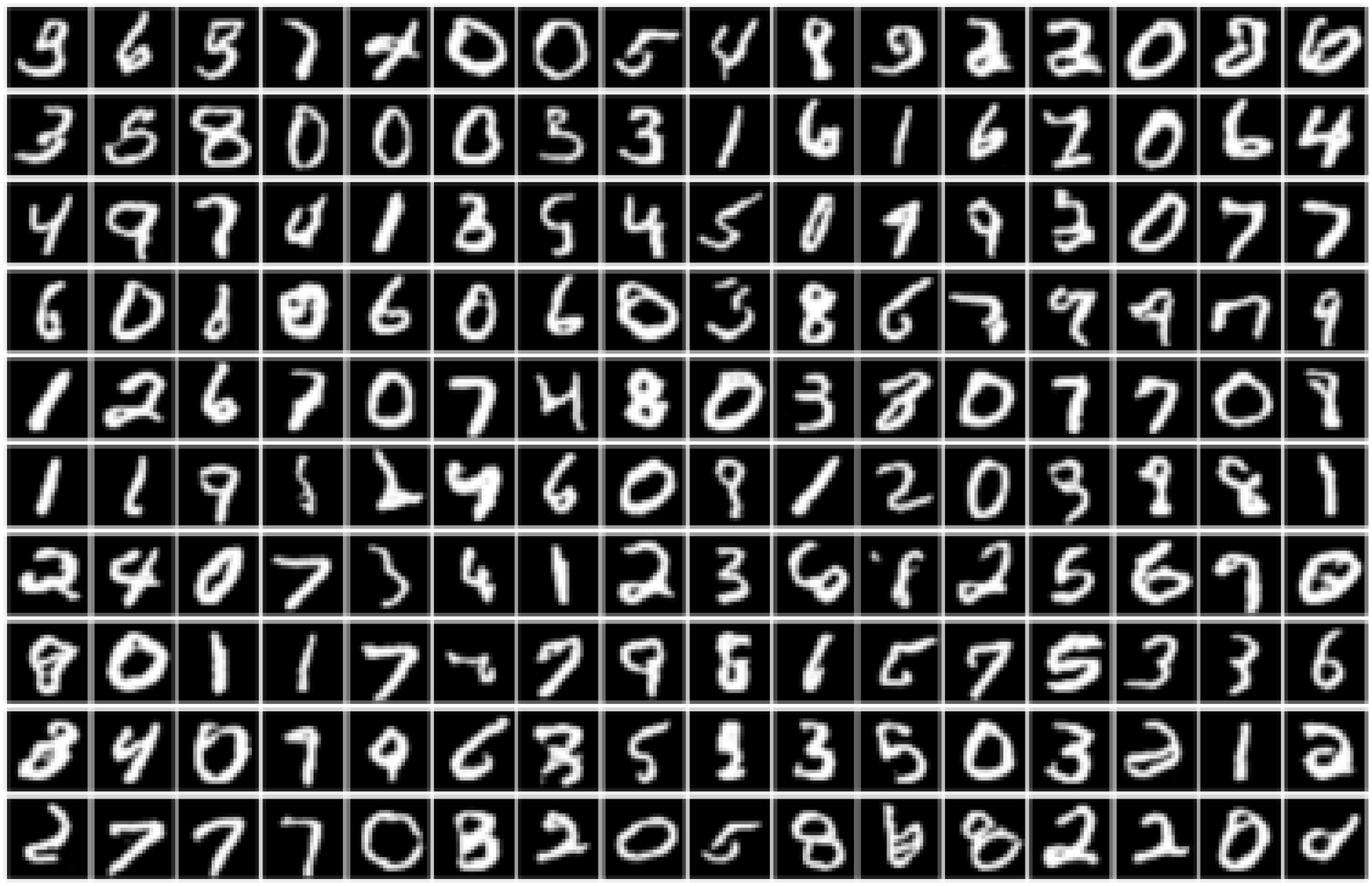}} \vskip1mm {\small
Fig.\,3\quad The generated images using ResGAN generator for the MNIST dataset.}}}

\begin{center} {\centering
\vbox{\centerline{\includegraphics[width=8cm,bb=0 -1 907 589]
{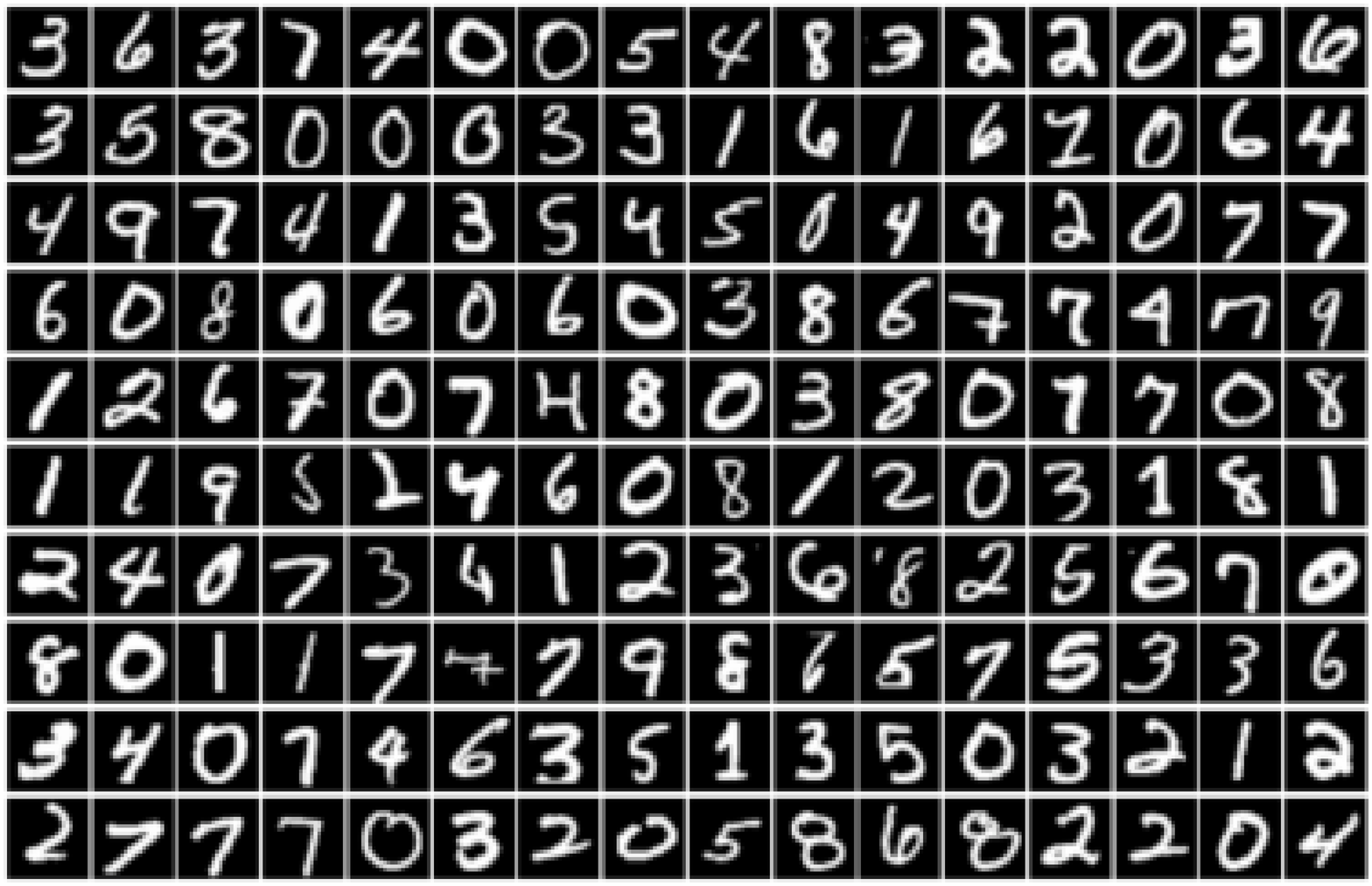}} \vskip1mm {\small
Fig.\,4\quad The realistic images selected from MNIST dataset as supervised attributes.}}}
\end{center}
\end{center}

\begin{center} {\centering
\vbox{\centerline{\includegraphics[width=8cm,bb=0 -1 907 589]
{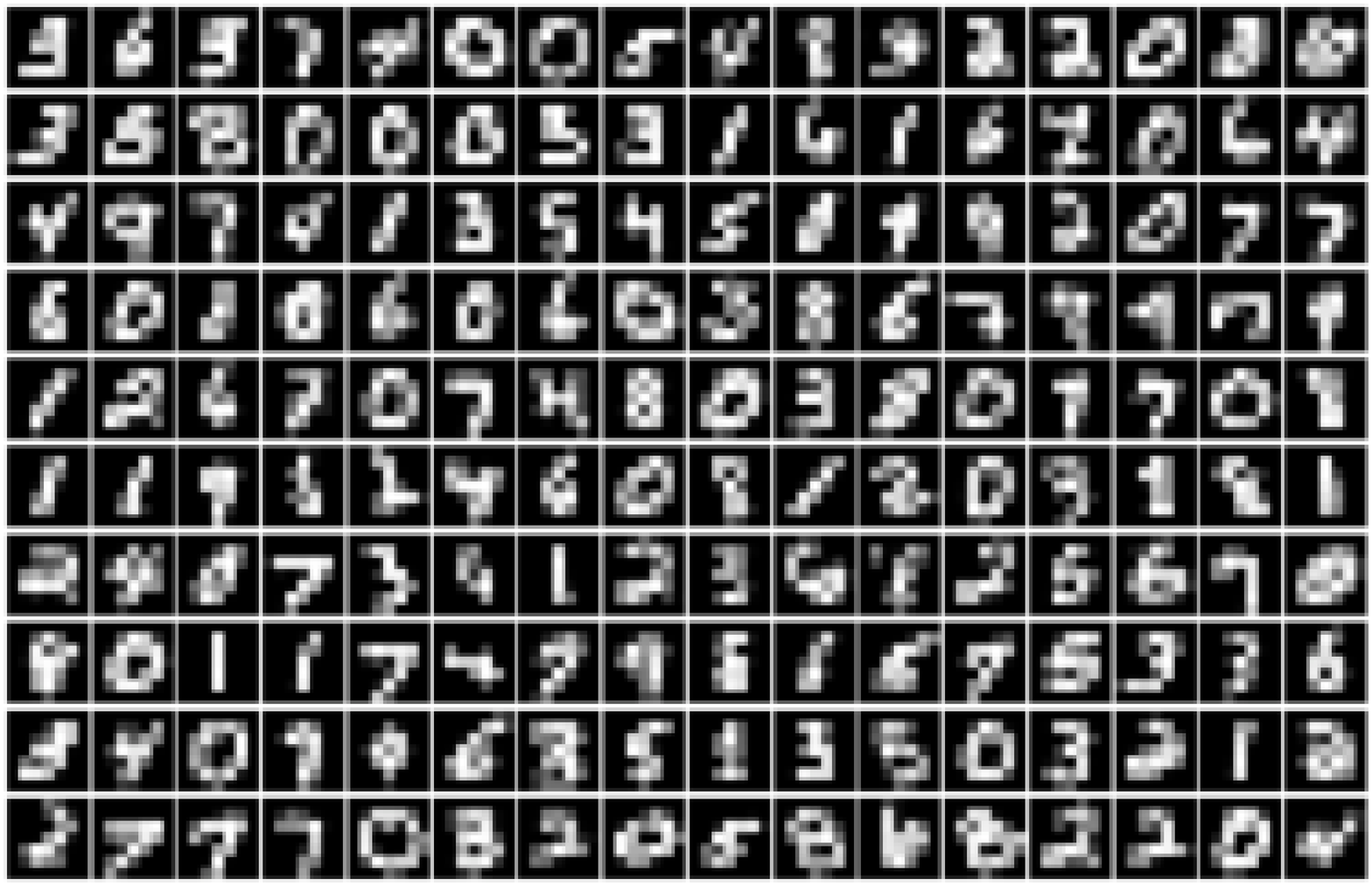}} \vskip1mm {\small
Fig.\,5\quad The corresponding coarse images as the inputs of the generative network.}}}
\end{center}

The losses and accuracies are illuminated in Fig. 6 and 7.
We can see the cross point of the losses of $G$ and $D$ in epoch 135.
It means that the balance is reached that indicates the best solution for both $G$ and $D$.
The iteration is not stop at the balance point and we set this configuration to obtain more generative samples and measurements. 

\begin{center} {\centering
\vbox{\centerline{\includegraphics[width=7cm,bb=0 -1 615 499]
{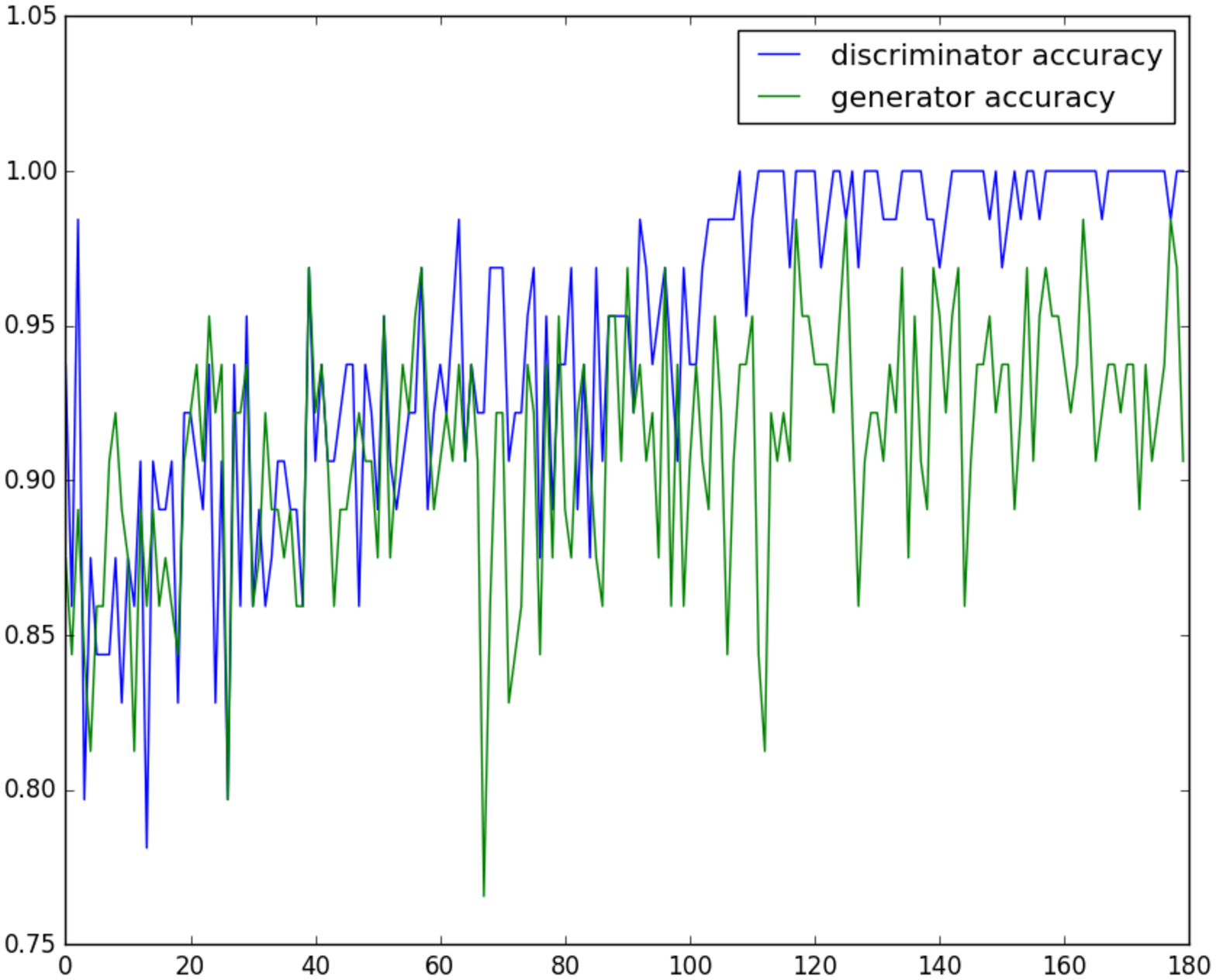}} \vskip1mm {\small
Fig.\,6\quad The curves of the classifier accuracies by the iterations of ResGAN model.}}}
\end{center}

\begin{center} {\centering
\vbox{\centerline{\includegraphics[width=7cm,bb=0 -1 607 499]
{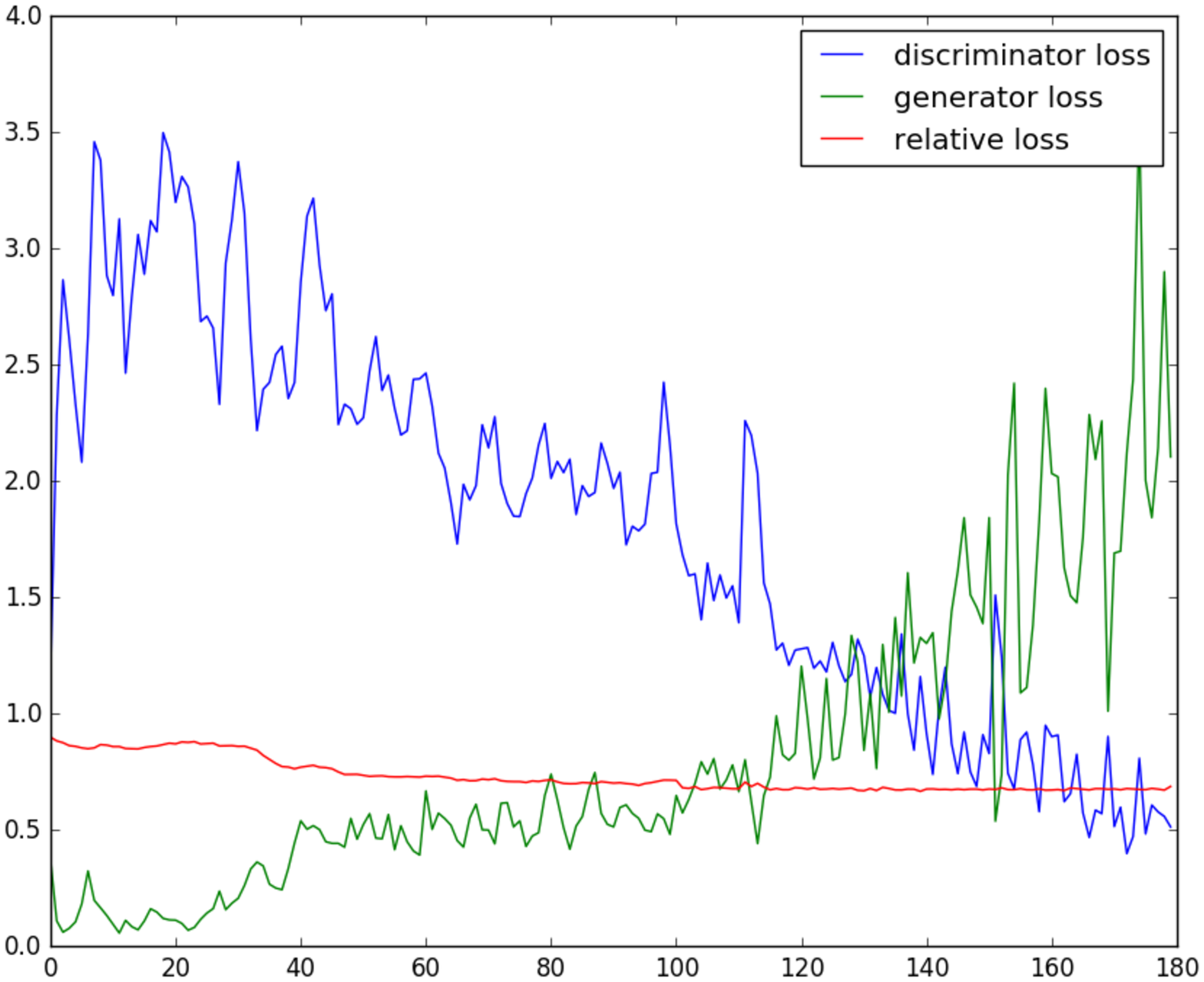}} \vskip1mm {\small
Fig.\,7\quad The curves of the logistic losses by the iterations of ResGAN model.}}}
\end{center}

We also test the GAN models on CIFAR10 and CIFAR100.
Both of the datasets contain various images that belong to several chosen classes,
and the main difference is the class number.
In Fig.8, it shows that the generated images have lots of details exactly similar to the realistic examples in Fig. 9.
And the input image shown in Fig. 10 are extremely coarse that the discriminated features are eliminated,
but the coarse spatial features are maintained. The ResGAN use these features by the directly connections in the Resnet structure. 

\begin{center} {\centering
\vbox{\centerline{\includegraphics[width=7.5cm,bb=0 -1 907 589]
{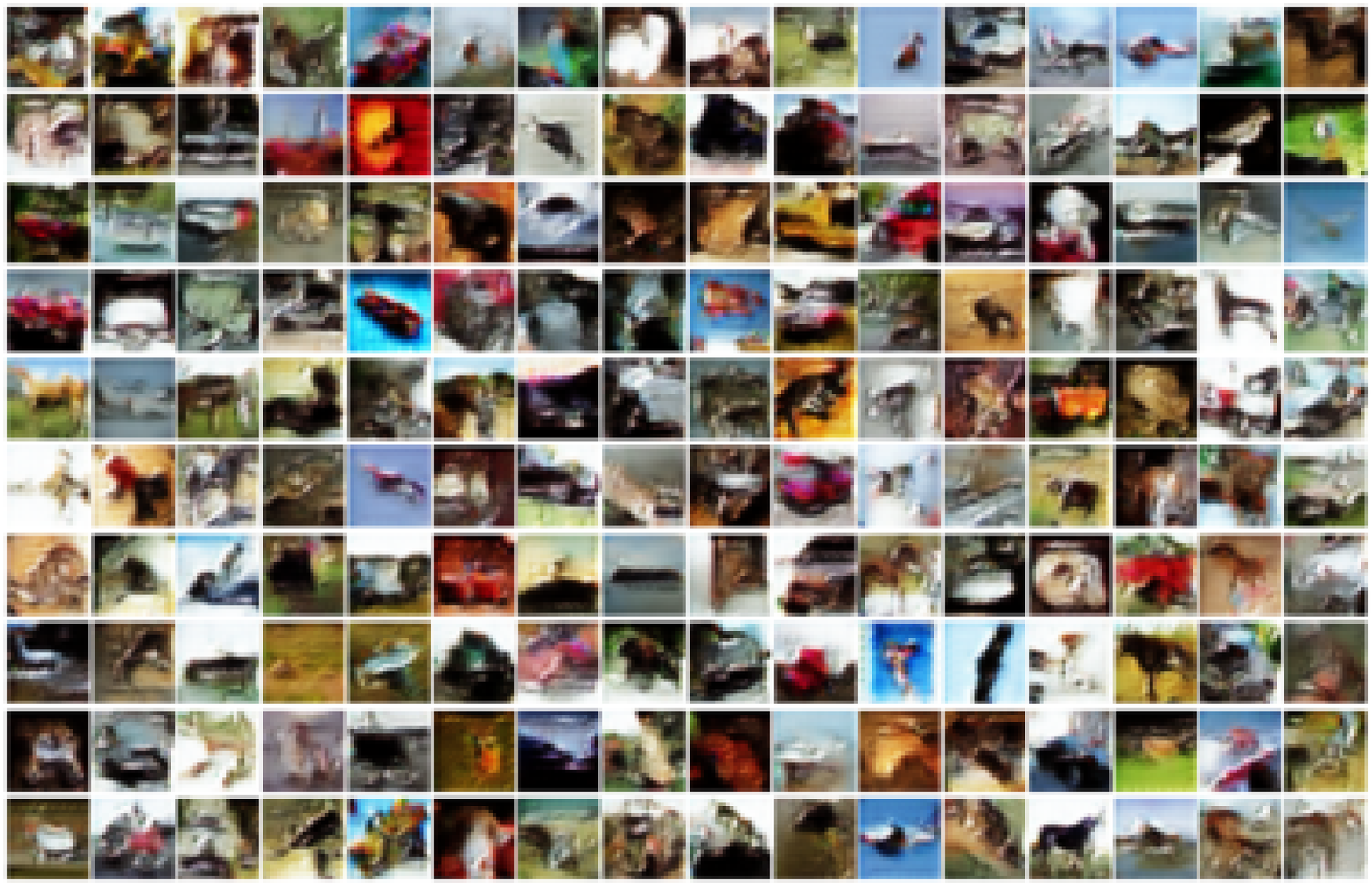}} \vskip1mm {\small
Fig.\,8\quad The generated images using ResGAN generator for the CIFAR10 dataset.}}}
\end{center}

\begin{center} {\centering
\vbox{\centerline{\includegraphics[width=7.5cm,bb=0 -1 907 589]
{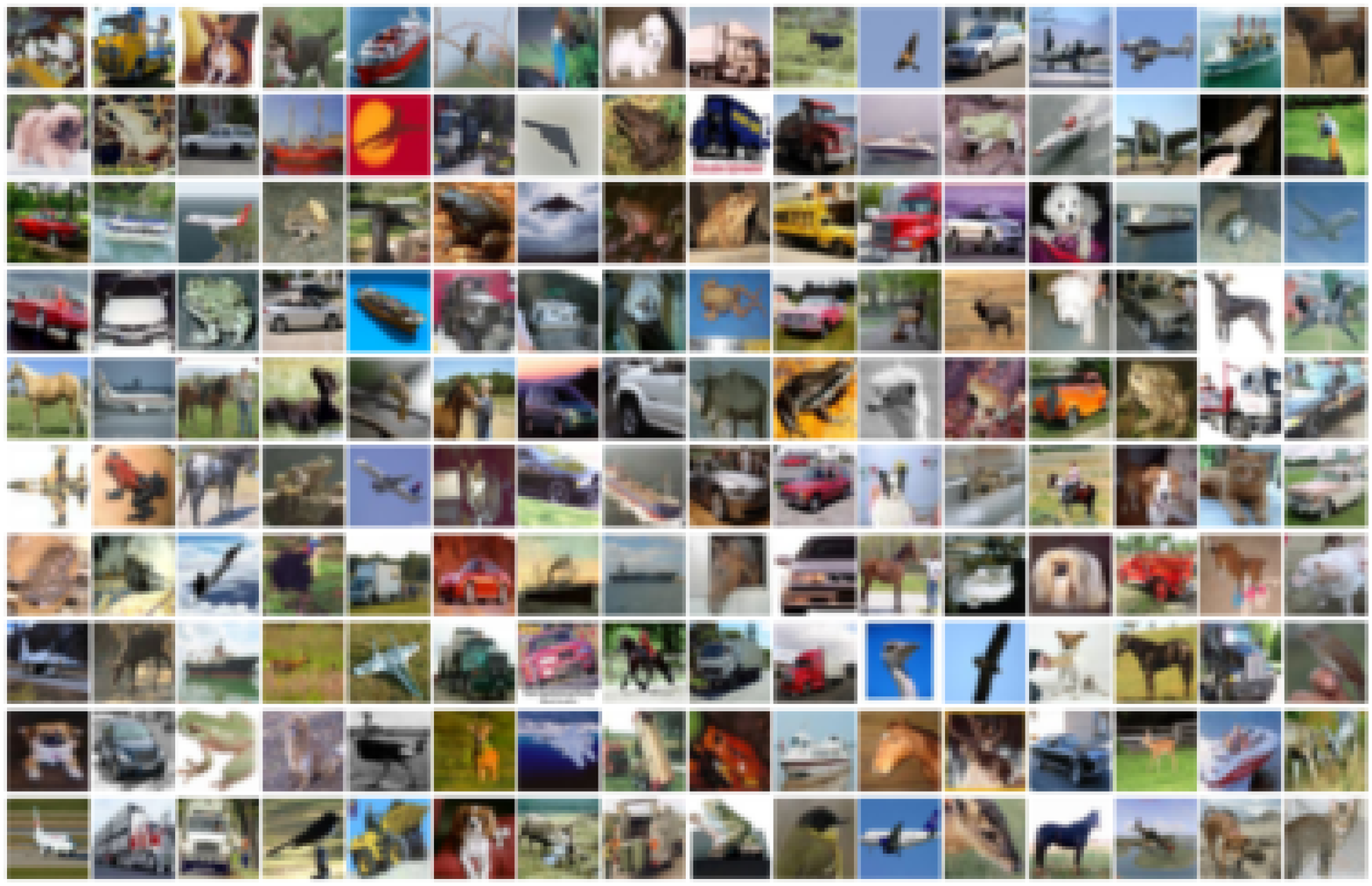}} \vskip1mm {\small
Fig.\,9\quad The realistic images selected from CIFAR10 dataset as supervised attributes.}}}
\end{center}

\begin{center} {\centering
\vbox{\centerline{\includegraphics[width=7.5cm,bb=0 -1 907 589]
{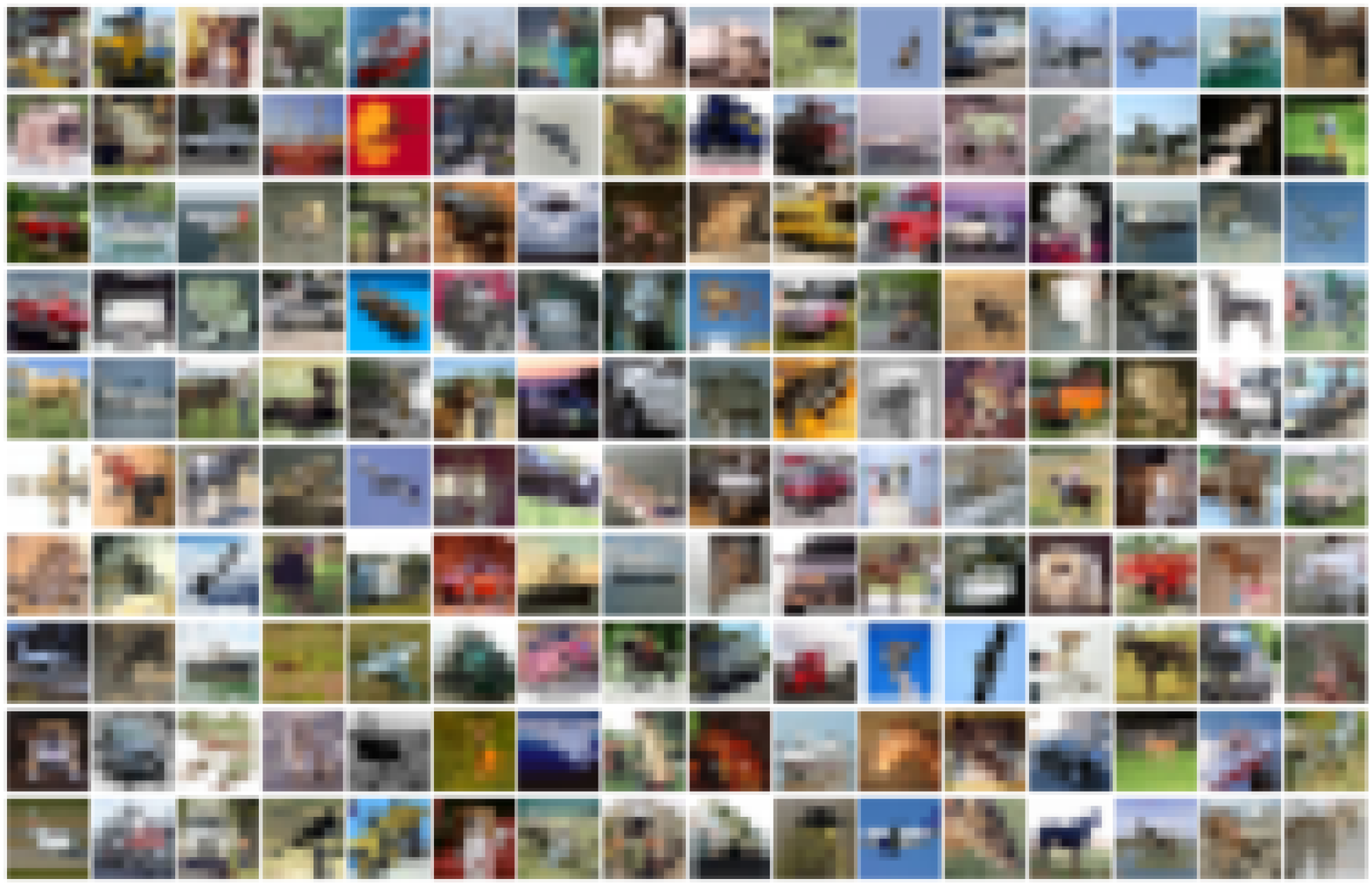}} \vskip1mm {\small
Fig.\,10\quad The corresponding coarse images as the inputs of the generative network.}}}
\end{center}

The testing results of CIFAR100 are shown in Fig. 11 and 12.
Compared to CIFAR10, the classes of CIFAR100 has reached 100, 
and it means that the intrinsic distributions of each classes of the dataset are simpler than CIFAR10.
It can be seen that the results in Fig. 11 show more discriminated details than Fig. 8. 
It is implied that the detailed classes each with simpler distribution are more easily to learn and generate,
especially compared to these coarse classes with complex distribution.

\begin{center} {\centering
\vbox{\centerline{\includegraphics[width=7.5cm,bb=0 -1 1135 1129]
{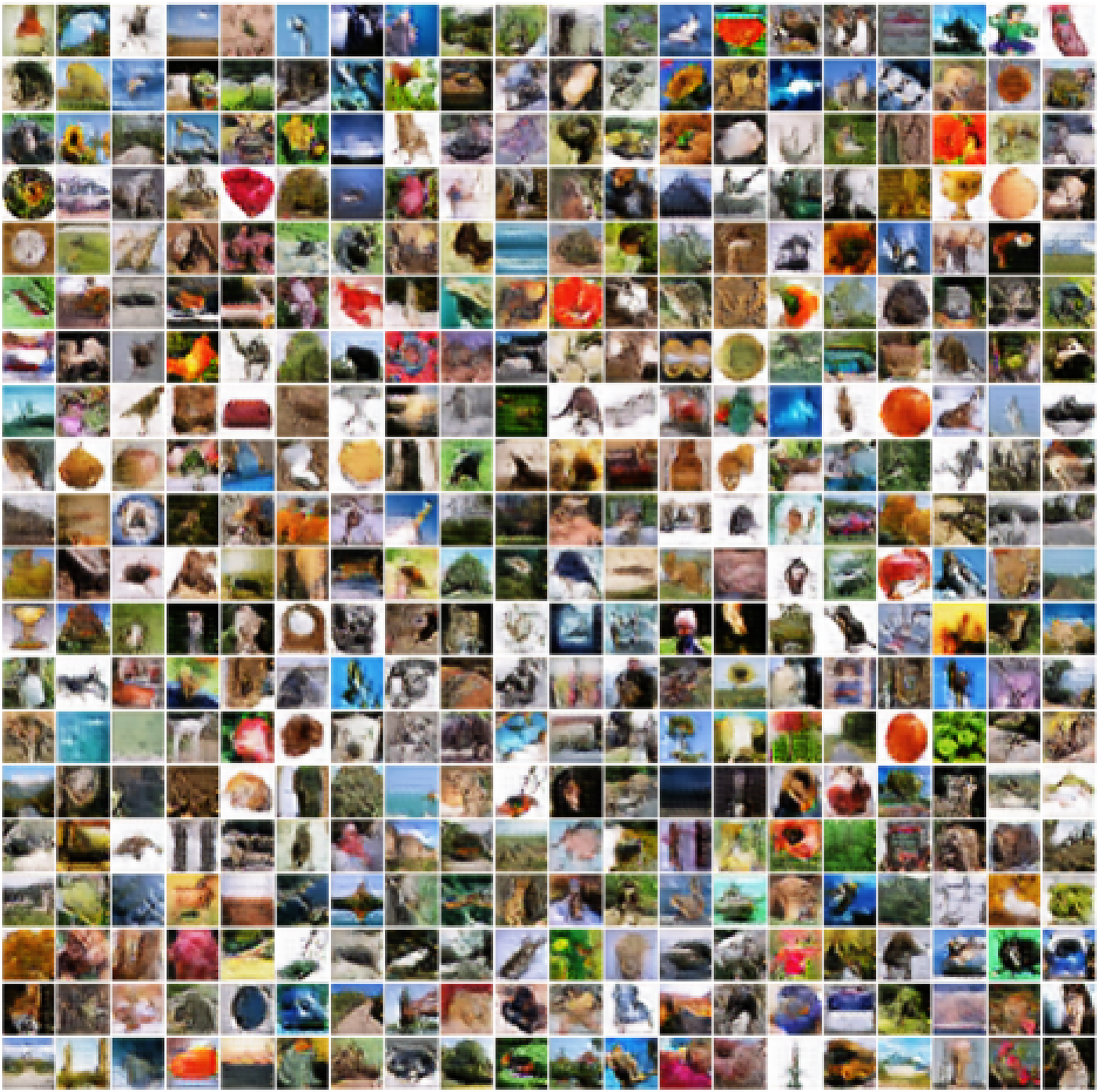}} \vskip1mm {\small
Fig.\,11\quad The generated images using ResGAN generator for the CIFAR100 dataset.}}}
\end{center}

\begin{center} {\centering
\vbox{\centerline{\includegraphics[width=7.5cm,bb=0 -1 1135 1129]
{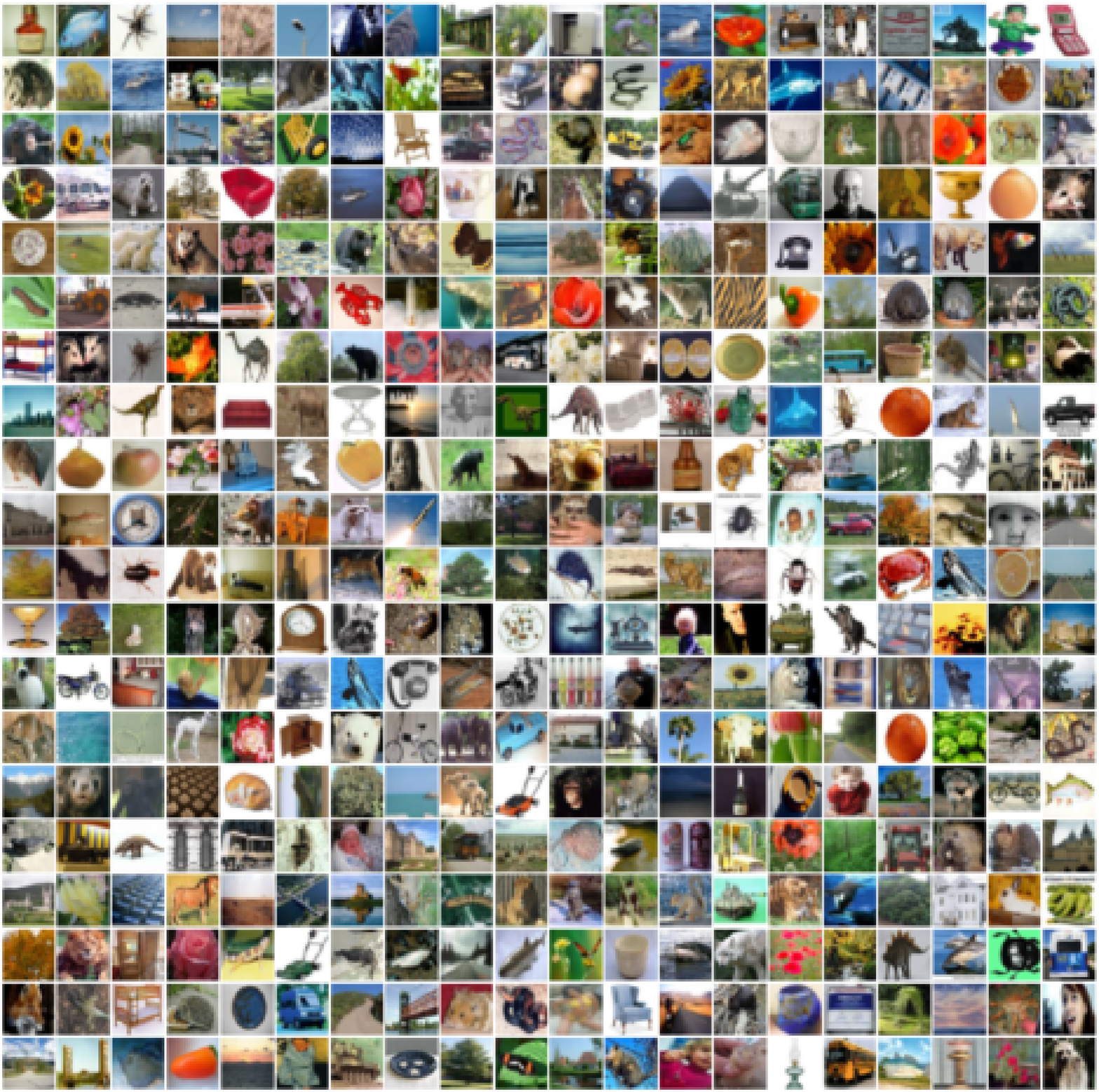}} \vskip1mm {\small
Fig.\,12\quad The realistic images selected from CIFAR100 dataset as supervised attributes.}}}
\end{center}

\begin{center}
\vbox{\centering{\small
Table 1 \quad The performances illuminated as ``losses/accuracies'' of the compared GAN models with the datasets. (
epoch=220)} \vskip2mm\renewcommand
    {\baselinestretch}{1.25}
    {\footnotesize\centerline{\tabcolsep=10pt\begin{tabular}
    {ccccc}
\toprule
{\sl CS}      & MNIST     & CIFAR10       & CIFAR100   & CelebA \\
\hline
GAN           & 3.54/.406     & 5.43/.423     & 9.53/.212     & 16.23/.093 \\
DCGAN         & 2.78/.815     & 4.66/.831     & 7.94/.519     & 13.45/.157 \\
WGAN        & 2.50/.870     & 3.79/.726     & 7.98/.384     & 13.72/.139 \\
CGAN          & 2.04/.912     & 3.58/.785     & 7.23/.473     & 12.44/.141 \\
ResGAN          & 1.98/.938     & 2.97/.794     & 6.55/.612     & 10.57/.237 \\
\bottomrule
\end{tabular}}}}
\end{center}

To evaluate the accuracies of different GAN modules, we record the losses/accuracies of 1000 iterative epochs and show the average measurements as in Tab. 1. 
Tab. 1 illuminates that according to the state-of-art GANs the generative performances to the datasets.
It denotes the ResGAN has the highest performance than the other models, and the second is CGAN that slightly better than the DCGAN and WGAN.
The loss function of WGAN is sharper and simpler than logistic function but not efficient for learning weights from large and complex dataset. 
The worst is GAN as the generator only has shallow network.
These measurements verified that the generation implemented by our scheme have excellent accuracy and loss performance. 

From extremely coarse face to rebuild discriminative examples is challenge since lots of various and precise features on the face region.
Thus, the generative model should benefit from a large dataset in order to deal with these situations. 
CelebA is one of the famous face dataset with 200599 faces and 40 attributes of each face that can be aliened to supervised labels for the extended discriminator.
The whole dataset is trained to formulate a face restore model.
The results shown in Fig. 7-9 verify that the ResGAN are suitable for the face restore task.
It can be seen in Fig. 13 the coarse faces is hard to discriminate even by human,
and as shown in Fig. 14 the ResGAN generate realistic faces in most of the test examples,
though some hardest examples are failed. 

\begin{center} {\centering
\vbox{\centerline{\includegraphics[width=8.5cm,bb=0 -1 912 574]
{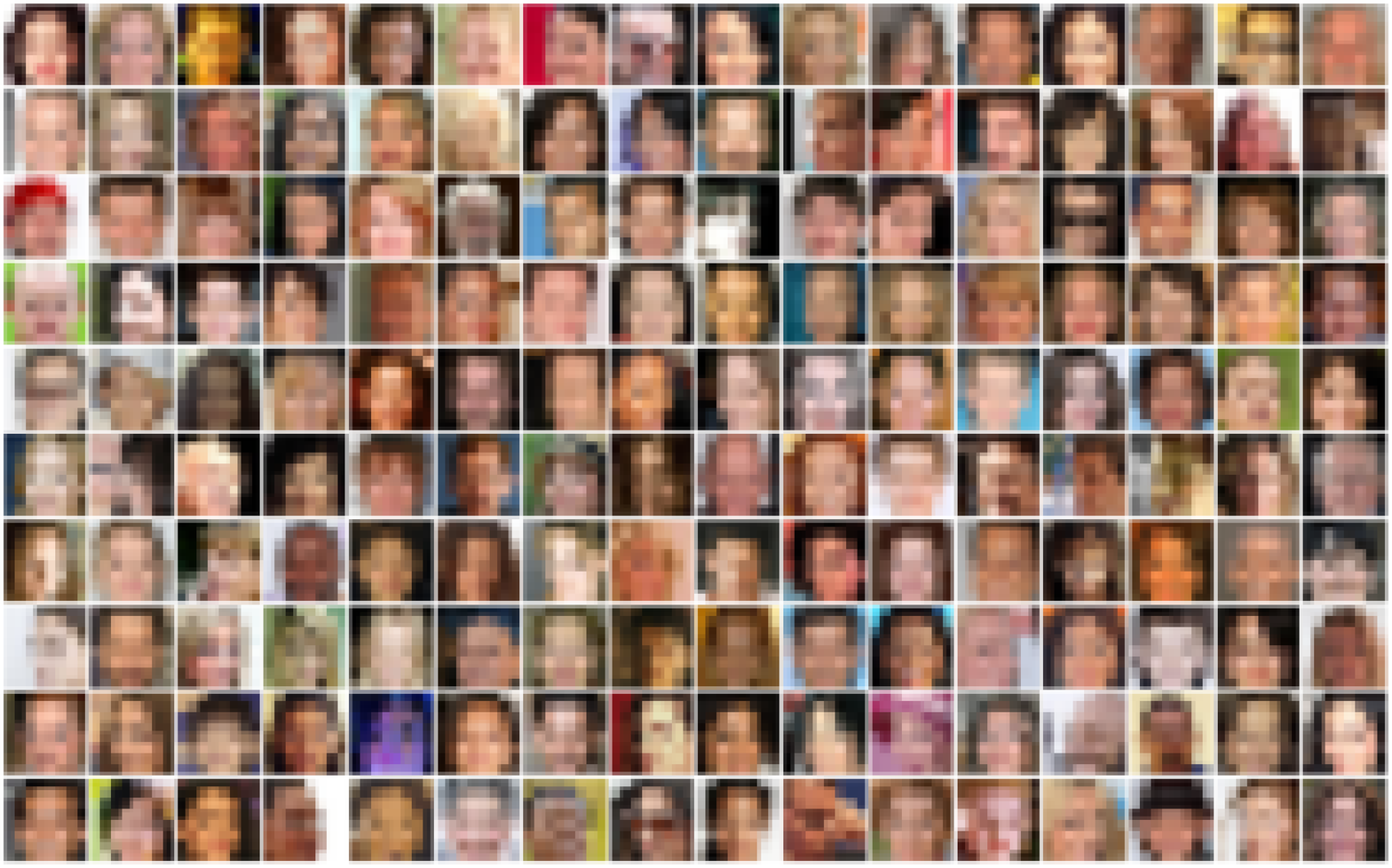}} \vskip1mm {\small
Fig.\,13\quad The corresponding coarse images as the inputs of the generative network.}}}
\end{center}

\begin{center} {\centering
\vbox{\centerline{\includegraphics[width=8.5cm,bb=0 -1 912 574]
{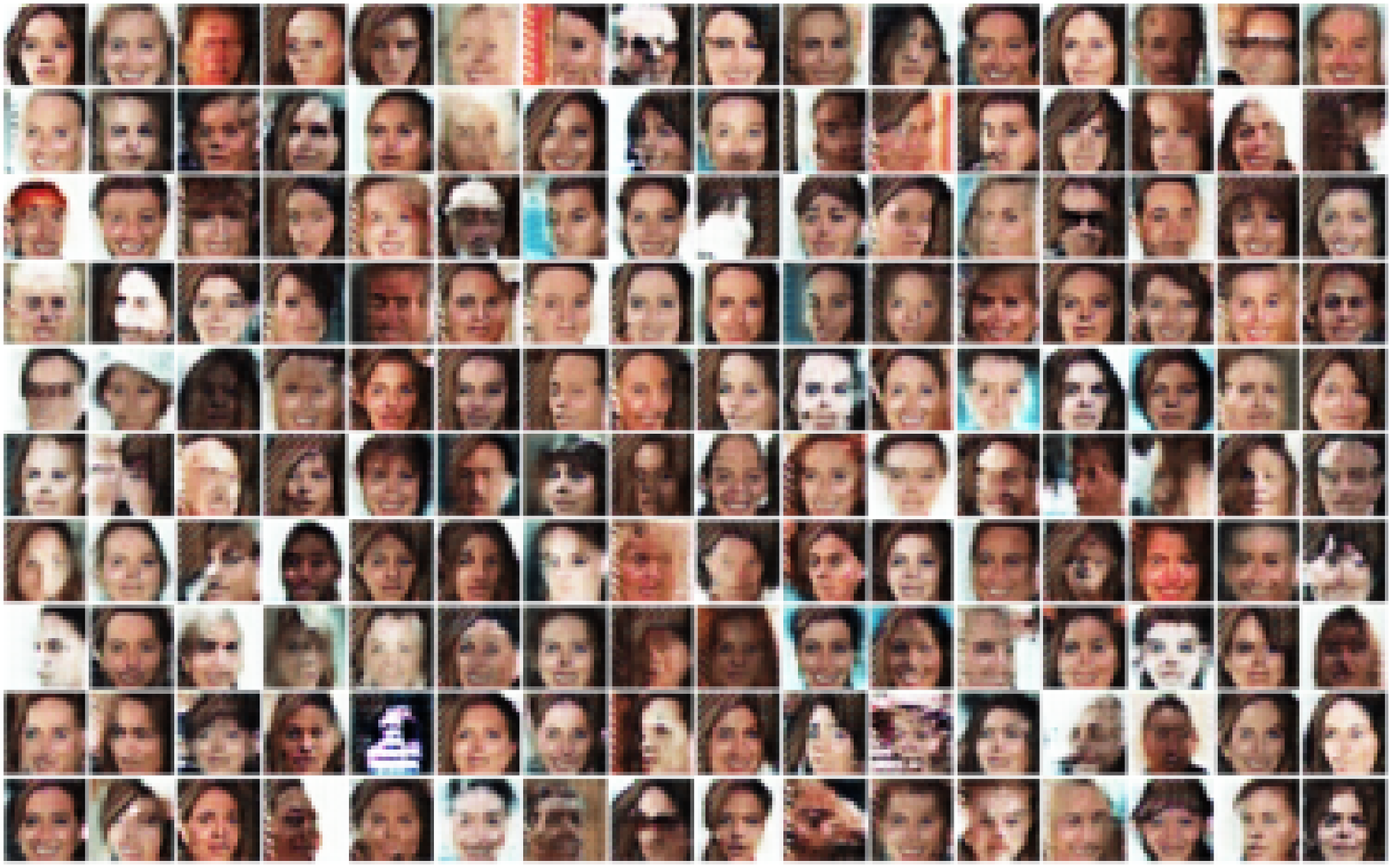}} \vskip1mm {\small
Fig.\,14\quad The generated images using ResGAN generator for the CelebA dataset.}}}
\end{center}

In addition, the corresponding real faces as background are illuminated in Fig. 15.
The ResGAN based on multiple convolution layers have ability to remember these faces of the train set,
and then it will be activated by the coarse faces as input of the generator.
It implies that by training the nodes in network are smart enough to know how to take part in the generation themselves.  

\begin{center} {\centering
\vbox{\centerline{\includegraphics[width=8.5cm,bb=0 -1 912 574]
{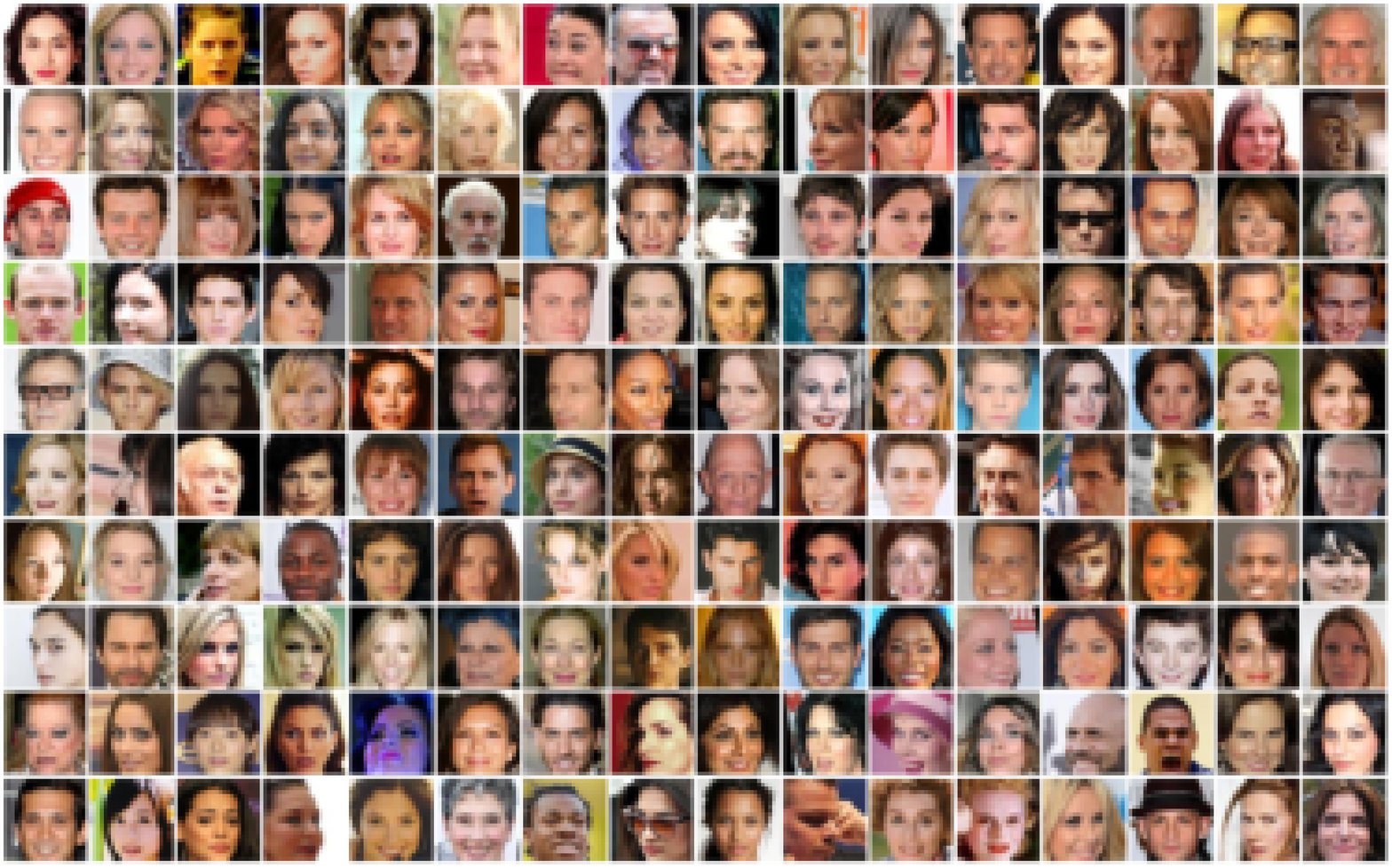}} \vskip1mm {\small
Fig.\,15\quad The realistic images selected from CelebA dataset as supervised attributes.}}}
\end{center}



\section{Conclusions}
The performances of image restores are improved by the proposed ResGAN model based on the deep convolution networks.
We test several famous datasets as MNIST, CIFAR10/100, CelebA and the results verified the lower losses and higher accuracies reached comparing to the other state-of-art GANs.
The main conclusion can be summarized including 2 aspects.
The one is that the classifier embedding can improve the GAN performance by guiding the processing of optimization directly.
And the second may lies in the Resnet structure is more stable and efficient for generative networks.
To further study the proposed ResGAN should be implemented to more configurations such as occlusion or noise exists. 
\acknowledgements
Support by National Natural Science Foundation of China:
Study on manifold vector field learning and anomaly discriminating of group interactive trajectory in vision scenes (61563025), 2016/01-2019/12.
And Yunnan Science and Technology Department of Science and Technology Project:
Study on semantic representation and discrimination of interactions based on sparse spatial-temporal BoW (2016FB109), 2016/10-2019/09.



\bibliographystyle{jaciiibibtex}
\bibliography{MyCollection}



\end{document}